\documentclass[a4paper]{article}

\usepackage{INTERSPEECH2022}
\usepackage{makecell}
\usepackage{array}
\usepackage[colorlinks=true, linkcolor=black, citecolor=black, urlcolor=black]{hyperref}

\title{Seewo's Submission to MLC-SLM:\\
      Lessons learned from Speech Reasoning Language Models}
\name{Bo Li, Chengben Xu, Wufeng Zhang}
\address{Seewo, Guangzhou, China}
\email{\{libo, xuchengben, zhangwufeng\}@cvte.com}

\begin{document}

\maketitle
\begin{abstract}

This paper presents Seewo's systems for both tracks of the Multilingual Conversational Speech Language Model Challenge (MLC-SLM), addressing automatic speech recognition (ASR) and speaker diarization with ASR (SD-ASR). We introduce a multi-stage training pipeline that explicitly enhances reasoning and self-correction in speech language models for ASR. Our approach combines curriculum learning for progressive capability acquisition, Chain-of-Thought data augmentation to foster intermediate reflection, and Reinforcement Learning with Verifiable Rewards (RLVR) to further refine self-correction. Our approach substantially outperforms the official baselines, achieving 11.57\% WER/CER (Track 1) and 17.67\% tcpWER/tcpCER (Track 2) on the evaluation set. Comprehensive ablation studies validate each component’s effectiveness.

\end{abstract}


\section{Introduction}

The MLC-SLM challenge focuses on multilingual conversational speech recognition and speaker diarization tasks, encompassing 11 languages: English (en), French (fr), German (de), Italian (it), Portuguese (pt), Spanish (es), Japanese (jp), Korean (ko), Russian (ru), Thai (th), and Vietnamese (vi). The English subset contains approximately 500 hours of recordings from diverse regions, including British, American, Australian, Indian, and Philippine English. Each of the remaining languages contributes around 100 hours, resulting in approximately 1,500 hours of multilingual conversational speech data. 

The challenge tasked participants with developing end-to-end speech language models for automatic speech recognition (ASR) and speaker diarization tasks, presenting significant technical challenges in multilingual processing, conversational speech understanding, and model optimization.

This paper presents the Seewo team's approach to the MLC-SLM challenge, with a particular focus on Track 1's multilingual ASR task. Our proposed system achieves substantial performance improvements over the baseline \cite{MLC-SLM-Baseline} through systematic architectural and training innovations. The key technical contributions include:

\begin{itemize}
\item A multi-stage pipeline that enhances the reasoning capabilities of speech language models through fine-tuning with curriculum learning strategy
\item An investigation of various reward functions for reinforcement learning (RL) based optimization of self-correction capabilities
\item An empirical study of contextual augmentation and decoding hyperparameters on ASR accuracy
\end{itemize}

\section{System overview}

This section details the core configuration of our system, including model architecture, data processing, computational resources and toolkits. All experiments used this configuration.

\subsection{Foundation models}

\begin{figure}[t]
  \centering
  \includegraphics[width=\linewidth]{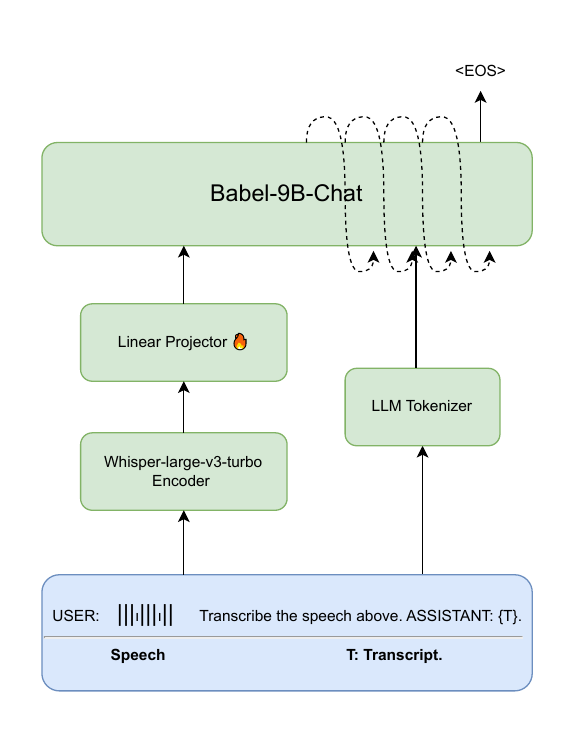}
  \caption{Model architecture referencing SLAM-ASR \cite{ma2024embarrassinglysimpleapproachllm}}
  \label{fig:model_architecture}
\end{figure}

Our speech language model architecture follows the SLAM-ASR framework design \cite{ma2024embarrassinglysimpleapproachllm}. SLAM-ASR provides a comprehensive evaluation of different encoder modules, including the Whisper family of models and other self-supervised encoders such as WavLM Large \cite{Chen_2022} and HuBERT X-Large \cite{hsu2021hubertselfsupervisedspeechrepresentation}. 

While these self-supervised models performed well in monolingual settings, their English-only pretraining limited their suitability for the multilingual task. Therefore, we adopted the Whisper large-v3-Turbo encoder \cite{radford2022whisper} for its robust multilingual capabilities and parameter efficiency.

We selected Babel-9B-Chat \cite{zhao2025babelopenmultilinguallarge} for the decoder component based on two key considerations. First, chat models typically outperform base pretrained models in instruction following \cite{ma2024embarrassinglysimpleapproachllm}. Second, Babel-9B-Chat's training data comprehensively covers all 11 languages required for the MLC-SLM challenge. As demonstrated in Table 7 of \cite{zhao2025babelopenmultilinguallarge}, it outperforms other 10B-size models on most multilingual benchmarks.

To bridge the encoder and decoder, we used a learnable projection module that maps encoder outputs to the decoder embedding space. The projector module employed a hierarchical architecture consisting of a convolutional layer followed by two linear layers, designed to downsample and align the encoder's output features with the decoder's embedding space. This 17.32M-parameter module effectively downsamples the speech features to 10Hz (one token per 100ms), aligning the temporal resolution with the LLM's input requirements. The complete model architecture is shown in Figure \ref{fig:model_architecture}.

\subsection{Training loss}

Our training pipeline employs a two-phase approach: Supervised Fine-Tuning (SFT) followed by Reinforcement Learning with Verifiable Rewards (RLVR). This design enables systematic enhancement of the model's ASR capabilities while maintaining stability.

In the SFT phase, we optimize the model using two objectives. First, we apply the standard causal language modeling loss to establish basic next-token prediction capabilities. Additionally, during the Chain-of-Thought (CoT) stage, we use a weighted loss variant that assigns different weights to various completion sections \cite{huertaenochian2024instructionfinetuningdoesprompt}, allowing the model to focus on critical transcription segments. With this approach, the model learns to transcribe the speech content after reasoning.

The RLVR phase adopts the Dr. GRPO \cite{liu2025understandingr1zeroliketrainingcritical} method, an optimized variant of the original GRPO framework \cite{shao2024deepseekmathpushinglimitsmathematical}. This approach improves the model performance by carefully balancing exploration of potential improvements with the maintenance of model stability through Kullback-Leibler (KL) divergence constraints with the reference model. The verifiable rewards ensure that the model's reasoning capabilities are enhanced in a controlled manner.

\subsection{Data and augmentation}

The ASR model is trained on the MLC-SLM challenge training set, as described in Section 1. 

The speaker embedding model for speaker diarization is trained on the MLC-SLM challenge training set and additional open-source datasets including CN-Celeb1, CN-Celeb2 \cite{li2020cn}, VoxBlink \cite{lin2023voxblinklargescalespeaker} and VoxBlink2 \cite{lin2024voxblink2100kspeakerrecognition}, to build a robust multilingual speaker embedding model.

For evaluation, we used the development set (approximately 4 hours per language) as our primary test set because the challenge did not provide an official evaluation set and limited the number of submissions. All ablation data reported in this paper are based on the development set.

We applied simple additive noise and reverberation augmentation techniques \cite{wang2023wespeaker} to enhance model robustness. For each training sample, data augmentation was applied with a probability of 0.7; when augmentation was performed, either additive noise or reverberation was randomly selected. Room Impulse Response (RIR) and noise data were randomly sampled from the SLR28 dataset \cite{7953152} and the MUSAN dataset \cite{musan2015}, respectively.

\subsection{Training toolkits and resources}
  Our training pipeline leverages several toolkits, including west \cite{west2024wenet}, transformers \cite{wolf-etal-2020-transformers}, accelerate \cite{accelerate}, peft \cite{peft} , and trl \cite{vonwerra2022trl}. All experiments were conducted on a cluster of 24 NVIDIA A800 GPUs, each equipped with 80GB of memory.

\section{Training pipeline of ASR system}

We adopt a multi-stage training pipeline where each stage builds upon the previous checkpoint to address specific challenges. While we present only the successful experiments here, our development process included numerous exploratory attempts that, despite not improving performance, provided valuable insights into the model's capabilities and limitations.

\subsection{Stage 1: Projector training}

The chat template is also following the SLAM-ASR \cite{ma2024embarrassinglysimpleapproachllm}: "USER: \texttt{<S>} \texttt{<P>} ASSISTANT: \texttt{<T>}", where \texttt{<S>} represents speech embedding, \texttt{<P>} represents the prompt, which is "Transcribe the speech above" in English for all data, and \texttt{<T>} represents the corresponding transcribed text.

We train the projector module (freezing all parameters except the projector module) in the first 2000 steps, checkpoint-2000 (ckpt-2000). At the end of this stage, the loss converges to around 0.35. This ensures the speech features are aligned with the decoder's embedding space and the LLM roughly follows the transcription instruction. 

\subsection{Stage 2: Special tokens adaptation}

Following Stage 1, we observed that the model occasionally deviated from the transcription task, generating conversational responses to the speech content rather than producing transcriptions. To address this instruction-following issue, we extended the special tokens to enhance the model's structured output capabilities, inspired by Qwen2.5 \cite{qwen2025qwen25technicalreport} and LaVIT \cite{jin2024unifiedlanguagevisionpretrainingllm}.

We extended Babel-9B-Chat's vocabulary with four categories of functional tokens:
\begin{itemize}
    \item \texttt{<LANG\_XX>}, Language-specific tokens for precise language identification across all 11 target languages
    \item \texttt{<speech>} \texttt{</speech>}, Speech segment delimiters to clearly demarcate speech input boundaries
    \item \texttt{<transcribe>} \texttt{</transcribe>}, Task control tokens to enforce transcription-focused output generation
    \item \texttt{<think>} \texttt{</think>}, Reasoning framework tokens to facilitate explicit intermediate reasoning
\end{itemize}

To accommodate these new tokens, we selectively unfroze the embedding layer and language model head, enabling the model to fine-tune the weights for token representations while maintaining the integrity of the pre-trained knowledge.

The enhanced prompt template follows the basic format "USER: \texttt{<S>} \texttt{<P>} ASSISTANT: \texttt{<T>}", with two key modifications from Stage 1:
\begin{itemize}
    \item \texttt{<S>}, Speech input is encapsulated within \texttt{<speech>} and \texttt{</speech>} tags to establish distinct input boundaries.
    \item \texttt{<T>}, Assistant outputs an explicit structured format as "\texttt{<LANG\_XX>} \texttt{<transcribe>} [transcription] \texttt{</transcribe>}".
\end{itemize}

This tokenized approach significantly improved the model's task adherence and output consistency, leading to a loss of approximately 0.2 after 4000 steps (ckpt-4000).

\subsection{Stage 3: LoRA training with multilingual prompts and context}

While Stage 2 improved instruction following, two challenges remained: (1) code-mixing between English and the target languages in transcriptions, likely due to English prompt interference, and (2) a training loss plateau at 0.2, indicating potential for further optimization.

To address these issues, we used LoRA \cite{hu2021loralowrankadaptationlarge} adaptation (with rank 16 and lora\_alpha 32) on both the LLM decoder and Whisper encoder. Additionally, we introduced language-specific prompts in the \texttt{<P>} section, which were "Transcribe the speech above" translated into each respective language.

This adaptation strategy proved effective in tackling the code-mixing issue, reducing the training loss to 0.15 after 7000 steps (ckpt-7000).

\subsection{Stage 4: SFT for reflection}

Despite improvements in previous stages, the model continued to exhibit typical ASR errors, such as substitutions, insertions, and deletions. Recent studies \cite{ma2025s2rteachingllmsselfverify} \cite{kumar2024traininglanguagemodelsselfcorrect} have shown that large language models can be trained to self-verify and self-correct. To achieve this, we first trained the model with Chain-of-Thought (CoT) data to explicitly reflect mistakes before outputting the transcription.

The CoT prompt template maintains the basic structure "USER: \texttt{<S>} \texttt{<P>} ASSISTANT: \texttt{<T>}" but introduces a reasoning part in \texttt{<T>} section, following the format: "\texttt{<LANG\_XX>} \texttt{<think>} The speech sounds like: $hypothesis_1$, but it might have some \texttt{[error\_details]}, let me correct it. \texttt{</think>} \texttt{<transcribe>} $hypothesis_2$ \texttt{</transcribe>}"

The $hypothesis_1$ is generated with the Stage 3 checkpoint (ckpt-7000) by running inference on the training data. For each hypothesis, the \texttt{[error\_details]} are generated by computing the WER/CER against the ground truth transcriptions, and specific error details are identified whenever the WER is non-zero, ensuring that at least one relevant error is included when present in the hypothesis. The ground-truth transcription is used as $hypothesis_2$.

We continued fine-tuning the model from ckpt-7000 of Stage 3. The resulted in a training loss of 0.12 after 9000 steps (ckpt-9000). However, despite the lower training loss, evaluation on the development set revealed a WER/CER of 35\%, which was significantly worse than that of ckpt-7000. Further analysis showed that the model often produced correct transcriptions in the $hypothesis_1$ section while generating incorrect final outputs in the \texttt{<transcribe>} tags. This indicates that the model failed to effectively transfer its intermediate reasoning to the final transcription output.

To understand this counterintuitive result, we analyzed the loss function in detail. We identified a fundamental issue with the standard causal modeling loss (Eq. \ref{eq:loss_c}):

\begin{equation}
  L_c = -\frac{1}{|I_c|} \sum_{i \in I_c} \log \mathbb{P}(t_i | t_{<i}; \theta)
  \label{eq:loss_c}
\end{equation}

We hypothesize that the standard loss is suboptimal for this task due to (1) completion length $|I_c|$ inversely affecting loss magnitude, and (2) the typically longer \texttt{<think>} section disproportionately influencing the loss. Recent studies \cite{shi2024instructiontuninglossinstructions,huertaenochian2024instructionfinetuningdoesprompt} have shown that the completion-to-prompt length ratio ($R_g$) significantly impacts model performance. In our experiments, this imbalance resulted in close loss values for stages 3 and 4, despite substantial differences in WER.

To address this, we implemented a modified version of the prompt loss token weighting (PLW) strategy \cite{huertaenochian2024instructionfinetuningdoesprompt}. Unlike the original approach, we applied $PLW < 1$ to the \texttt{<think>} section while maintaining full masking for the prompt. This modification focused the model's attention on the \texttt{<transcribe>} section through the following loss calculation:

\begin{equation}
  L = \frac{\displaystyle-\sum_{i=1}^{N} w_i \cdot \log p_i}{\displaystyle\sum_{i=1}^{N} w_i}
\end{equation}

where $w_i=1$ for the \texttt{<transcribe>} section and $w_i=PLW$ for the \texttt{<think>} section.

We resumed training from ckpt-9000 with a decaying PLW schedule: starting at 1.0 and reducing to 0.1 over the first 300 steps, then maintaining 0.1. This achieved a final loss of 0.1 after 10000 steps (ckpt-10000).

The PLW strategy successfully refocused the model's learning on the transcription segment, resulting in a substantial reduction in WER/CER, as summarized in Table~\ref{tab:wer_cer}.

\begin{table}[h]
  \caption{WER/CER and loss of the SFT stages}
  \label{tab:wer_cer}
  \centering
  \begin{tabular}{ r@{}l  r  r}
    \toprule
    \multicolumn{2}{c}{\textbf{Stage}} & 
    \multicolumn{1}{c}{\textbf{training loss}} &
    \multicolumn{1}{c}{\textbf{WER/CER}} \\
    \midrule
    ckpt & -2000 & 0.35 & 24.32\% \\
    ckpt & -4000 & 0.2 &  18.99\% \\
    ckpt & -7000 & 0.15 & 15.48\% \\
    ckpt & -9000 & 0.12 & 35.03\% \\
    ckpt & -10000 & \textbf{0.1} & \textbf{13.42\%} \\
  \end{tabular}
\end{table}

\subsection{Stage 5: RLVR for reflection}

While Stage 4 successfully established the CoT completion pattern, two critical issues remained: (1) the WER/CER performance still fell short of our best experimental results, and (2) the reasoning content in the \texttt{<think>} section lacked meaningful analysis. After looking into the model's outputs, we identified several systematic failure modes in the reasoning process:

\begin{itemize}
  \item Inaccurate error type identification in the initial hypothesis analysis
  \item Insufficient or missing error details in the reflection phase
  \item Inability to apply correct error details to final transcription, even when accurately identified
  \item Generation of non-existent errors (hallucinations) in the reasoning process
\end{itemize}

To address these issues, we implemented a series of reward functions to guide the model to generate more meaningful reasoning content using Dr.GRPO \cite{liu2025understandingr1zeroliketrainingcritical}, which addressed the length bias of original GRPO \cite{shao2024deepseekmathpushinglimitsmathematical}. In our experiments, we rolled out 4 samples for each group, with multinomial sampling and beam search decoding. The sampling temperature is set to 0.5 to control the diversity of the samples.

Here are the reward functions we implemented:

\begin{itemize}
  \item $RF_1$: verifies the structural correctness of the output with respect to \texttt{<think>} and \texttt{<transcribe>} tags.
  \item $RF_2$: evaluates the accuracy of $hypothesis_2$ with respect to the ground truth (scaled between 0 and 1).
  \item $RF_3$: verifies whether the error types identified in the reflection match the actual errors present in $hypothesis_1$.
  \item $RF_4$: verifies whether the specific error details described in the reflection correspond to the actual errors in $hypothesis_1$.
  \item $RF_5$: evaluates whether $hypothesis_2$ achieves a lower WER/CER than $hypothesis_1$ with respect to the ground truth, rewarding only if $hypothesis_2$ is more accurate.
\end{itemize}

$RF_1$ and $RF_2$ constrain the basic performance of the model, including the format of output and the accuracy of the transcription. $RF_3$ and $RF_4$ are used to guide the model to generate more accurate reflection content. $RF_5$ is used to guide the model to generate more accurate transcriptions.

We conducted experiments with different combinations of these reward functions, with all post-training experiments based on ckpt-10000. Each reward score was scaled to 0 to 1.0, and the final advantage was calculated by the sum of the reward scores, weighted 0.8 for the $RF_2$ and 0.5 for the others. Table \ref{tab:reward_func} shows the WER/CER results for different combinations of reward functions, compared with the best model in SFT stages.

\begin{table}[h]
  \caption{Results of different combinations of reward functions}
  \label{tab:reward_func}
  \centering
  \begin{tabular}{l|c}
    \toprule
    Reward Functions & WER/CER \\
    \midrule
    ckpt-10000 (without RL) & 13.42\% \\
    $RF_1$+$RF_2$ & 13.15\% \\
    $RF_1$+$RF_2$+$RF_3$ & 15.44\% \\
    $RF_1$+$RF_2$+$RF_3$+$RF_4$ & 16.12\% \\
    $RF_1$+$RF_2$+$RF_5$ & \textbf{12.73\%} \\
    $RF_1$+$RF_5$ & 28.29\% \\
    $RF_2$+$RF_5$ & 20.75\% \\
  \end{tabular}

\end{table}

The combination of $RF_1$, $RF_2$, and $RF_5$ yielded the best performance among all reward function configurations in our experiments. This approach was inspired by SCoRe~\cite{kumar2024traininglanguagemodelsselfcorrect}, which uses a two-step correction mechanism to enhance final model performance. In our experiments, the reward associated with $RF_5$ consistently converged above 0.8, indicating that the model developed strong self-correction capabilities. However, we also observed a phenomenon of reward hacking, where $hypothesis_1$ progressively collapsed and became increasingly shorter after several hundred training steps. To mitigate this issue, we introduced an additional constraint requiring $hypothesis_1$ to be sufficiently similar to $hypothesis_2$ in order to receive rewards.

For $RF_3$, we observed that the error type identification was getting more accurate during training, and the reward of $RF_3$ converged above 0.6; however, this did not translate into improved WER/CER. For the $RF_4$, we found that the model was not able to generate the correct error details, the reward of this function converged around 0.5. We conjecture that the reward signal was too sparse to guide the model in generating the correct error details. 

In our experiments, the outcome-based reward functions were found to be an effective strategy for training models to perform self-correction, even though the intermediate reasoning steps did not demonstrate a direct causal influence on the final output. This observation aligns with recent findings in the literature \cite{havrilla2024glorewhenwhereimprove} \cite{besta2025reasoninglanguagemodelsblueprint} that suggest the process-based reward functions might be less stability and need more refinement. By the end of the challenge we had not found a good way to improve the reflection ability of the model, which would be a future work.

Furthermore, we found it was important to acquire the capability step by step with a curriculum learning strategy, the model's low level capability must be established first. In the attempt of $RF_2$+$RF_5$, the \texttt{<think>} structure format was collapsed quickly, and the rewards of $RF_2$ and $RF_5$ remained low. The GRPO reinforcement learning process attempted to obtain a better sampling signal, based on the premise that better sampling results exist in a group. \cite{shao2024deepseekmathpushinglimitsmathematical}

At the end of our post-training experiments, we achieved a WER/CER of 12.73\% with $RF_1$+$RF_2$+$RF_5$, with a checkpoint called ckpt-12500.

\section{Additional Experiments}

\subsection{Effect of Conversational Context on ASR}

Inspired by GEC-RAG \cite{robatian2025gecragimprovinggenerativeerror}, which demonstrated that contextual information can enhance error correction in post-processing, we explored the impact of conversational context on ASR model performance in the MLC-SLM challenge. Given the conversational and long-form nature of the audio, we hypothesized that incorporating preceding utterances could improve the model's self-correction ability.
During training, we augmented each sample by appending the transcriptions from the previous two utterances in the same conversation to the user prompt. With the new prompt, we fine-tuned the model from ckpt-7000 of Stage 3 to ckpt-9000. At the test time, the previous two utterances were used as context as well. The results are summarized in Table~\ref{tab:train_and_decode_with_context}.

\begin{table}[h]
  \caption{Results of training and decoding with context}
  \label{tab:train_and_decode_with_context}
  \centering
  \begin{tabular}{ r@{}l  r  r}
    \toprule
    \multicolumn{2}{c}{\textbf{Stage}} & 
    \multicolumn{1}{c}{\textbf{training loss}} &
    \multicolumn{1}{c}{\textbf{WER/CER}} \\
    \midrule
    ckpt & -7000 & 0.15 & 15.48\% \\
    ckpt & -9000-context (with context) & \textbf{0.14} &  \textbf{14.30\%} \\
  \end{tabular}
\end{table}

It should be noted that, in this experiment, the contextual transcriptions used during evaluation were taken from the ground truth of the development set, rather than generated by the model itself. As a result, the reported results represent an upper bound on the potential benefit of context, rather than the true end-to-end performance. Due to the inefficiency of our inference pipeline for sequential utterance processing, we were unable to conduct large-scale, fully automatic experiments. Improving the inference efficiency and conducting stricter end-to-end evaluations will be the focus of our future work.

\subsection{Different decoding methods and hyperparameters}

We explored different decoding methods and hyperparameters using the best performance model, including different beam size, different decoding length, different decoding method, etc. All experiments were conducted with the best checkpoint ckpt-12500. The results are shown in table \ref{tab:different_decoding_methods}. The best result in our experiments is 12.73\% WER/CER with Beam+Sampling method, beam size 8, new max length 180.

\begin{table}[!htbp]
  \footnotesize
  \caption{Results of different decoding methods}
  \label{tab:different_decoding_methods}
  \centering
  \begin{tabular}{p{2.5cm} c c c}
    \toprule
    \multicolumn{1}{c}{\textbf{method}} & 
    \multicolumn{1}{c}{\textbf{beam size}} & 
    \multicolumn{1}{c}{\textbf{new max length}} & 
    \multicolumn{1}{c}{\textbf{WER/CER}} \\
    \midrule
    multinomial sampling & - & 150 & 16.72\% \\
    beam search & 10 & 150 & 14.04\% \\
    Beam+Sampling & 10 & 150 & 13.90\% \\
    Beam+Sampling & 12 & 150 & 14.26\% \\
    Beam+Sampling & 15 & 150 & 16.39\% \\
    Beam+Sampling & 8 & 150 & 13.55\% \\
    Beam+Sampling & 5 & 150 & 13.68\% \\
    Beam+Sampling & 2 & 150 & 13.61\% \\
    Beam+Sampling & 8 & 180 & \textbf{12.73\%} \\
  \end{tabular}
\end{table}

Contrary to our initial expectations, increasing the beam size did not yield consistent performance improvements, revealing a non-monotonic relationship between beam size and model accuracy.

\section{Speaker diarization pipeline}

\begin{figure}[t]
  \centering
  \includegraphics[height=9.0cm]{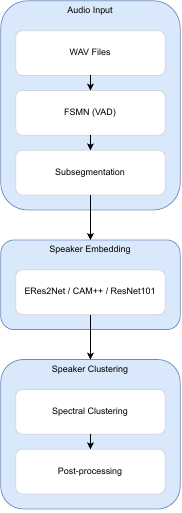}
  \caption{Pipeline of speaker diarization}
  \label{fig:sd}
\end{figure}

Our best speaker diarization system uses the 3D-Speaker-Toolkit \cite{chen20243d}, following the standard speaker diarization recipe without overlap detection. The system architecture comprises three key components: (1) a deep FSMN-based \cite{zhang2018deep} voice activity detection (VAD) module for precise speech segment identification, (2) a ResNet-101 speaker embedding model from Wespeaker \cite{wang2024advancing} for discriminative speaker representation, and (3) a spectral clustering module \cite{ning2006spectral} for speaker grouping. The complete pipeline is illustrated in Figure \ref{fig:sd}.

To optimize system performance, we conducted comprehensive ablation studies evaluating multiple speaker embedding models. Our experiments included the Wespeaker ResNet-101 \cite{wang2024advancing}, along with pre-trained models Eres2net \cite{chen2023enhancedres2netlocalglobal} and CAM++ \cite{campp} from ModelScope \cite{modelscope}. The ResNet-101 model was specifically trained on a diverse corpus comprising CnCeleb1, CnCeleb2 \cite{li2020cn}, VoxBlink \cite{lin2023voxblinklargescalespeaker}, and VoxBlink2 \cite{lin2024voxblink2100kspeakerrecognition} datasets.

A critical optimization involved adjusting the VAD module's \texttt{max\_single\_segment\_time} parameter. The default 60-second setting in 3D-Speaker-Toolkit exceeds the Whisper encoder's maximum length, degrading ASR performance. Furthermore, our training data analysis revealed that utterances typically fall under 10 seconds. Therefore, we systematically evaluated different combinations of speaker embedding models and segment durations. Table \ref{tab:sd_results} presents the performance metrics, demonstrating that the ResNet-101 model with an 8-second segment duration achieves the best performance, yielding 16.78\% DER and 18.03\% tcpWER.

\begin{table}
  \small
  \caption{Analysis of multiple speaker embedding models with different segment time configurations}
  \label{tab:sd_results}
  \centering
  \begin{tabular}{ r c c c }
    \toprule
    \textbf{Embedding Model} & 
    \textbf{Max Seg.Time(s)} & 
    \textbf{DER} &
    \textbf{tcpWER} \\
    \midrule
    CAM++      & 60 & 16.71\% & 24.64\% \\
    CAM++      & 8  & 17.28\% & 19.32\% \\
    Eres2net   & 8  & 17.22\% & 19.27\% \\
    ResNet101  & 8  & \textbf{16.78\%} & \textbf{18.03\%} \\
    \bottomrule
  \end{tabular}
\end{table}

\section{Conclusions}

We presented a multi-stage training pipeline for multilingual conversational ASR, integrating LoRA, Chain-of-Thought data augmentation, and RLVR-based self-correction. Our approach achieves substantial improvements over the official challenge baselines. Table \ref{tab:summary_results} and \ref{tab:summary_results_track2} summarize the results of our main submission to the evaluation set, demonstrating the effectiveness of each component. Despite these advances, challenges remain in fully automating context-aware inference and improving the granularity of self-correction. Future work will focus on enhancing inference efficiency, developing more sophisticated reward functions for reasoning, and extending our methods to broader real-world multilingual and conversational scenarios.

\begin{table}[h]
  \caption{Results of development and evaluation sets of track 1}
  \label{tab:summary_results}
  \centering
  \begin{tabular}{l c r r}
    \toprule
    \multicolumn{1}{c}{\textbf{}} &
    \multicolumn{1}{c}{\textbf{}} &
    \multicolumn{2}{r}{\textbf{\makecell{WER/CER}}} \\
    \multicolumn{1}{c}{\textbf{checkpoints}} & 
    \multicolumn{1}{c}{\textbf{method}} & 
    \multicolumn{1}{c}{\textbf{dev set}} &
    \multicolumn{1}{c}{\textbf{test set}} \\
    \midrule
    \makecell{baseline \cite{MLC-SLM-Baseline}} & \makecell{-} & \makecell{21.49\%} & \makecell{20.17\%} \\
    \makecell{ckpt-4000} & \makecell{Special token SFT} & \makecell{18.99\%} & \makecell{20.29\%} \\
    \makecell{ckpt-7000} & \makecell{LoRA SFT} & \makecell{15.48\%} & \makecell{13.86\%} \\
    \makecell{ckpt-10000} & \makecell{CoT SFT} & \makecell{13.42\%} & \makecell{12.26\%} \\
    \makecell{ckpt-12500-A} & \makecell{RLVF($RF_1$+$RF_2$)} & \makecell{13.15\%} & \makecell{11.79\%} \\
    \makecell{ckpt-12500-B} & \makecell{RLVF($RF_1$+$RF_2$+$RF_5$)} & \makecell{\textbf{12.73\%}} & \makecell{\textbf{11.57\%}} \\
  \end{tabular}
\end{table}

\begin{table}[h]
  \caption{Results of development and evaluation sets of track 2}
  \label{tab:summary_results_track2}
  \centering
  \begin{tabular}{c c c c}
    \toprule
    \multicolumn{1}{c}{\textbf{}} &
    \multicolumn{1}{c}{\textbf{}} &
    \multicolumn{2}{r}{\textbf{\makecell{tcpWER/tcpCER}}} \\
    \multicolumn{1}{c}{\textbf{checkpoints}} & 
    \multicolumn{1}{c}{\textbf{Max Seg.Time(s)}} & 
    \multicolumn{1}{c}{\textbf{dev set}} &
    \multicolumn{1}{c}{\textbf{test set}} \\
    \midrule
    \makecell{baseline \cite{MLC-SLM-Baseline}} & \makecell{-} & \makecell{76.12\%} & \makecell{60.39\%} \\
    \makecell{CAM++} & \makecell{60} & \makecell{24.64\%} & \makecell{22.39\%} \\
    \makecell{CAM++} & \makecell{8} & \makecell{19.32\%} & \makecell{18.34\%} \\
    \makecell{ResNet101} & \makecell{8} & \makecell{\textbf{18.03\%}} & \makecell{\textbf{17.67\%}} \\
  \end{tabular}
\end{table}

\clearpage



\begin{thebibliography}{10}
\providecommand{\url}[1]{#1}
\csname url@samestyle\endcsname
\providecommand{\newblock}{\relax}
\providecommand{\bibinfo}[2]{#2}
\providecommand{\BIBentrySTDinterwordspacing}{\spaceskip=0pt\relax}
\providecommand{\BIBentryALTinterwordstretchfactor}{4}
\providecommand{\BIBentryALTinterwordspacing}{\spaceskip=\fontdimen2\font plus
\BIBentryALTinterwordstretchfactor\fontdimen3\font minus \fontdimen4\font\relax}
\providecommand{\BIBforeignlanguage}[2]{{%
\expandafter\ifx\csname l@#1\endcsname\relax
\typeout{** WARNING: IEEEtran.bst: No hyphenation pattern has been}%
\typeout{** loaded for the language `#1'. Using the pattern for}%
\typeout{** the default language instead.}%
\else
\language=\csname l@#1\endcsname
\fi
#2}}
\providecommand{\BIBdecl}{\relax}
\BIBdecl

\bibitem{MLC-SLM-Baseline}
M.~Shen, ``Mlc-slm-baseline,'' \url{https://github.com/mubingshen/MLC-SLM-Baseline}, 2025, accessed: 2024-03-20.

\bibitem{ma2024embarrassinglysimpleapproachllm}
Z.~Ma, G.~Yang, Y.~Yang, Z.~Gao, J.~Wang, Z.~Du, F.~Yu, Q.~Chen, S.~Zheng, S.~Zhang, and X.~Chen, ``An embarrassingly simple approach for llm with strong asr capacity,'' 2024.

\bibitem{Chen_2022}
S.~Chen, C.~Wang, Z.~Chen, Y.~Wu, S.~Liu, Z.~Chen, J.~Li, N.~Kanda, T.~Yoshioka, X.~Xiao, J.~Wu, L.~Zhou, S.~Ren, Y.~Qian, Y.~Qian, J.~Wu, M.~Zeng, X.~Yu, and F.~Wei, ``Wavlm: Large-scale self-supervised pre-training for full stack speech processing,'' \emph{IEEE Journal of Selected Topics in Signal Processing}, vol.~16, no.~6, p. 1505–1518, Oct. 2022.

\bibitem{hsu2021hubertselfsupervisedspeechrepresentation}
W.-N. Hsu, B.~Bolte, Y.-H.~H. Tsai, K.~Lakhotia, R.~Salakhutdinov, and A.~Mohamed, ``Hubert: Self-supervised speech representation learning by masked prediction of hidden units,'' 2021.

\bibitem{radford2022whisper}
A.~Radford, J.~W. Kim, T.~Xu, G.~Brockman, C.~McLeavey, and I.~Sutskever, ``Robust speech recognition via large-scale weak supervision,'' 2022.

\bibitem{zhao2025babelopenmultilinguallarge}
Y.~Zhao, C.~Liu, Y.~Deng, J.~Ying, M.~Aljunied, Z.~Li, L.~Bing, H.~P. Chan, Y.~Rong, D.~Zhao, and W.~Zhang, ``Babel: Open multilingual large language models serving over 90

\bibitem{huertaenochian2024instructionfinetuningdoesprompt}
M.~Huerta-Enochian and S.~Y. Ko, ``Instruction fine-tuning: Does prompt loss matter?'' 2024.

\bibitem{liu2025understandingr1zeroliketrainingcritical}
Z.~Liu, C.~Chen, W.~Li, P.~Qi, T.~Pang, C.~Du, W.~S. Lee, and M.~Lin, ``Understanding r1-zero-like training: A critical perspective,'' 2025.

\bibitem{shao2024deepseekmathpushinglimitsmathematical}
Z.~Shao, P.~Wang, Q.~Zhu, R.~Xu, J.~Song, X.~Bi, H.~Zhang, M.~Zhang, Y.~K. Li, Y.~Wu, and D.~Guo, ``Deepseekmath: Pushing the limits of mathematical reasoning in open language models,'' 2024.

\bibitem{li2020cn}
L.~Li, R.~Liu, J.~Kang, Y.~Fan, H.~Cui, Y.~Cai, R.~Vipperla, T.~F. Zheng, and D.~Wang, ``Cn-celeb: multi-genre speaker recognition,'' 2020.

\bibitem{lin2023voxblinklargescalespeaker}
Y.~Lin, X.~Qin, G.~Zhao, M.~Cheng, N.~Jiang, H.~Wu, and M.~Li, ``Voxblink: A large scale speaker verification dataset on camera,'' 2023.

\bibitem{lin2024voxblink2100kspeakerrecognition}
Y.~Lin, M.~Cheng, F.~Zhang, Y.~Gao, S.~Zhang, and M.~Li, ``Voxblink2: A 100k+ speaker recognition corpus and the open-set speaker-identification benchmark,'' 2024.

\bibitem{wang2023wespeaker}
H.~Wang, C.~Liang, S.~Wang, Z.~Chen, B.~Zhang, X.~Xiang, Y.~Deng, and Y.~Qian, ``Wespeaker: A research and production oriented speaker embedding learning toolkit,'' in \emph{IEEE International Conference on Acoustics, Speech and Signal Processing (ICASSP)}.\hskip 1em plus 0.5em minus 0.4em\relax IEEE, 2023, pp. 1--5.

\bibitem{7953152}
T.~Ko, V.~Peddinti, D.~Povey, M.~L. Seltzer, and S.~Khudanpur, ``A study on data augmentation of reverberant speech for robust speech recognition,'' in \emph{2017 IEEE International Conference on Acoustics, Speech and Signal Processing (ICASSP)}, 2017, pp. 5220--5224.

\bibitem{musan2015}
D.~Snyder, G.~Chen, and D.~Povey, ``{MUSAN}: {A} {M}usic, {S}peech, and {N}oise {C}orpus,'' 2015, arXiv:1510.08484v1.

\bibitem{west2024wenet}
W.~Team, ``We speech transcript, llm based speech recognition/transcript in 300 lines of code,'' \url{https://github.com/wenet-e2e/west}, 2024, accessed: 2024-03-20.

\bibitem{wolf-etal-2020-transformers}
T.~Wolf, L.~Debut, V.~Sanh, J.~Chaumond, C.~Delangue, A.~Moi, P.~Cistac, T.~Rault, R.~Louf, M.~Funtowicz, J.~Davison, S.~Shleifer, P.~von Platen, C.~Ma, Y.~Jernite, J.~Plu, C.~Xu, T.~L. Scao, S.~Gugger, M.~Drame, Q.~Lhoest, and A.~M. Rush, ``Transformers: State-of-the-art natural language processing,'' in \emph{Proceedings of the 2020 Conference on Empirical Methods in Natural Language Processing: System Demonstrations}.\hskip 1em plus 0.5em minus 0.4em\relax Online: Association for Computational Linguistics, Oct. 2020, pp. 38--45.

\bibitem{accelerate}
S.~Gugger, L.~Debut, T.~Wolf, P.~Schmid, Z.~Mueller, S.~Mangrulkar, M.~Sun, and B.~Bossan, ``Accelerate: Training and inference at scale made simple, efficient and adaptable.'' \url{https://github.com/huggingface/accelerate}, 2022.

\bibitem{peft}
S.~Mangrulkar, S.~Gugger, L.~Debut, Y.~Belkada, S.~Paul, and B.~Bossan, ``Peft: State-of-the-art parameter-efficient fine-tuning methods,'' \url{https://github.com/huggingface/peft}, 2022.

\bibitem{vonwerra2022trl}
L.~von Werra, Y.~Belkada, L.~Tunstall, E.~Beeching, T.~Thrush, N.~Lambert, S.~Huang, K.~Rasul, and Q.~Gallouédec, ``Trl: Transformer reinforcement learning,'' \url{https://github.com/huggingface/trl}, 2020.

\bibitem{qwen2025qwen25technicalreport}
Qwen, :, A.~Yang, B.~Yang, B.~Zhang, B.~Hui, B.~Zheng, B.~Yu, C.~Li, D.~Liu, F.~Huang, H.~Wei, H.~Lin, J.~Yang, J.~Tu, J.~Zhang, J.~Yang, J.~Yang, J.~Zhou, J.~Lin, K.~Dang, K.~Lu, K.~Bao, K.~Yang, L.~Yu, M.~Li, M.~Xue, P.~Zhang, Q.~Zhu, R.~Men, R.~Lin, T.~Li, T.~Tang, T.~Xia, X.~Ren, X.~Ren, Y.~Fan, Y.~Su, Y.~Zhang, Y.~Wan, Y.~Liu, Z.~Cui, Z.~Zhang, and Z.~Qiu, ``Qwen2.5 technical report,'' 2025.

\bibitem{jin2024unifiedlanguagevisionpretrainingllm}
Y.~Jin, K.~Xu, K.~Xu, L.~Chen, C.~Liao, J.~Tan, Q.~Huang, B.~Chen, C.~Lei, A.~Liu, C.~Song, X.~Lei, D.~Zhang, W.~Ou, K.~Gai, and Y.~Mu, ``Unified language-vision pretraining in llm with dynamic discrete visual tokenization,'' 2024.

\bibitem{hu2021loralowrankadaptationlarge}
E.~J. Hu, Y.~Shen, P.~Wallis, Z.~Allen-Zhu, Y.~Li, S.~Wang, L.~Wang, and W.~Chen, ``Lora: Low-rank adaptation of large language models,'' 2021.

\bibitem{ma2025s2rteachingllmsselfverify}
R.~Ma, P.~Wang, C.~Liu, X.~Liu, J.~Chen, B.~Zhang, X.~Zhou, N.~Du, and J.~Li, ``S$^2$r: Teaching llms to self-verify and self-correct via reinforcement learning,'' 2025.

\bibitem{kumar2024traininglanguagemodelsselfcorrect}
A.~Kumar, V.~Zhuang, R.~Agarwal, Y.~Su, J.~D. Co-Reyes, A.~Singh, K.~Baumli, S.~Iqbal, C.~Bishop, R.~Roelofs, L.~M. Zhang, K.~McKinney, D.~Shrivastava, C.~Paduraru, G.~Tucker, D.~Precup, F.~Behbahani, and A.~Faust, ``Training language models to self-correct via reinforcement learning,'' 2024.

\bibitem{shi2024instructiontuninglossinstructions}
Z.~Shi, A.~X. Yang, B.~Wu, L.~Aitchison, E.~Yilmaz, and A.~Lipani, ``Instruction tuning with loss over instructions,'' 2024.

\bibitem{havrilla2024glorewhenwhereimprove}
A.~Havrilla, S.~Raparthy, C.~Nalmpantis, J.~Dwivedi-Yu, M.~Zhuravinskyi, E.~Hambro, and R.~Raileanu, ``Glore: When, where, and how to improve llm reasoning via global and local refinements,'' 2024.

\bibitem{besta2025reasoninglanguagemodelsblueprint}
M.~Besta, J.~Barth, E.~Schreiber, A.~Kubicek, A.~Catarino, R.~Gerstenberger, P.~Nyczyk, P.~Iff, Y.~Li, S.~Houliston, T.~Sternal, M.~Copik, G.~Kwaśniewski, J.~Müller, Łukasz Flis, H.~Eberhard, Z.~Chen, H.~Niewiadomski, and T.~Hoefler, ``Reasoning language models: A blueprint,'' 2025.

\bibitem{robatian2025gecragimprovinggenerativeerror}
A.~Robatian, M.~Hajipour, M.~R. Peyghan, F.~Rajabi, S.~Amini, S.~Ghaemmaghami, and I.~Gholampour, ``Gec-rag: Improving generative error correction via retrieval-augmented generation for automatic speech recognition systems,'' 2025.

\bibitem{chen20243d}
Y.~Chen, S.~Zheng, H.~Wang, L.~Cheng \emph{et~al.}, ``3d-speaker-toolkit: An open source toolkit for multi-modal speaker verification and diarization,'' 2025.

\bibitem{zhang2018deep}
S.~Zhang, M.~Lei, Z.~Yan, and L.~Dai, ``Deep-fsmn for large vocabulary continuous speech recognition,'' in \emph{2018 IEEE International Conference on Acoustics, Speech and Signal Processing (ICASSP)}.\hskip 1em plus 0.5em minus 0.4em\relax IEEE, 2018, pp. 5869--5873.

\bibitem{wang2024advancing}
S.~Wang, Z.~Chen, B.~Han, H.~Wang, C.~Liang, B.~Zhang, X.~Xiang, W.~Ding, J.~Rohdin, A.~Silnova \emph{et~al.}, ``Advancing speaker embedding learning: Wespeaker toolkit for research and production,'' \emph{Speech Communication}, vol. 162, p. 103104, 2024.

\bibitem{ning2006spectral}
H.~Ning, M.~Liu, H.~Tang, and T.~S. Huang, ``A spectral clustering approach to speaker diarization.'' in \emph{Interspeech}.\hskip 1em plus 0.5em minus 0.4em\relax Citeseer, 2006.

\bibitem{chen2023enhancedres2netlocalglobal}
Y.~Chen, S.~Zheng, H.~Wang, L.~Cheng, Q.~Chen, and J.~Qi, ``An enhanced res2net with local and global feature fusion for speaker verification,'' 2023.

\bibitem{campp}
H.~Wang, S.~Zheng, Y.~Chen, L.~Cheng, and Q.~Chen, ``Cam++: A fast and efficient network for speaker verification using context-aware masking,'' \emph{arXiv preprint arXiv:2303.00332}.

\bibitem{modelscope}
T.~M. Team, ``Modelscope: bring the notion of model-as-a-service to life.'' \url{https://github.com/modelscope/modelscope}, 2023.

\end{thebibliography}
\end{document}